\documentclass[conference,a4paper]{IEEEtran}
\usepackage{cite}

\ifCLASSINFOpdf
  \usepackage[pdftex]{graphicx}
\else
\fi
\usepackage[cmex10]{amsmath}
\usepackage{bm}
\usepackage{nicefrac}
\usepackage{algorithm}
\usepackage{algpseudocode}
\algnewcommand\algorithmicforeach{\textbf{for each}}
\algdef{S}[FOR]{ForEach}[1]{\algorithmicforeach\ #1\ \algorithmicdo}
\DeclareRobustCommand*{\IEEEauthorrefmark}[1]{\raisebox{0pt}[0pt][0pt]{\textsuperscript{\footnotesize #1}}}

\renewcommand{\baselinestretch}{1.15}
\begin{document}
%
\title{Meta-Learning in Spiking Neural Networks with Reward-Modulated STDP}

\author{\IEEEauthorblockN{
Arsham Gholamzadeh Khoee\IEEEauthorrefmark{1},   
Alireza Javaheri\IEEEauthorrefmark{2},   
Saeed Reza Kheradpisheh\IEEEauthorrefmark{2},    
Mohammad Ganjtabesh\IEEEauthorrefmark{1}      
}                                     
\IEEEauthorblockA{\IEEEauthorrefmark{1}
Department of Computer Science, School of Mathematics, Statistics, and Computer Science, \\
College of Science, University of Tehran, Tehran, Iran}
\IEEEauthorblockA{\IEEEauthorrefmark{2}
Department of Computer Sciences, Shahid Beheshti University, G.C., Tehran, Iran}
}



\maketitle

\begin{abstract}
The human brain constantly learns and rapidly adapts to new situations by integrating acquired knowledge and experiences into memory. Developing this capability in machine learning models is considered an important goal of AI research since deep neural networks perform poorly when there is limited data or when they need to adapt quickly to new unseen tasks.
Meta-learning models are proposed to facilitate quick learning in low-data regimes by employing absorbed information from the past. Although some models have recently been introduced that reached high-performance levels, they are not biologically plausible. We have proposed a bio-plausible meta-learning model inspired by the hippocampus and the prefrontal cortex using spiking neural networks with a reward-based learning system. Our proposed model includes a memory designed to prevent catastrophic forgetting, a phenomenon that occurs when meta-learning models forget what they have learned as soon as the new task begins. Also, our new model can easily be applied to spike-based neuromorphic devices and enables fast learning in neuromorphic hardware. The final analysis will discuss the implications and predictions of the model for solving few-shot classification tasks. In solving these tasks, our model has demonstrated the ability to compete with the existing state-of-the-art meta-learning techniques.
\end{abstract}

{\smallskip \keywords Meta-Learning, Few-Shot Learning, Learning to Learn, Spiking Neurons, STDP, Reward-Modulated STDP, PFC, Hippocampus.}

%
\IEEEpeerreviewmaketitle

\vspace{7pt}
\section{Introduction}
\label{sec:intro}

Today’s machine learning and deep learning models excel at solving single tasks; however, they struggle when training data is insufficient or they have to adapt to changing tasks~\cite{schmidhuber2015deep}. Meta-learning is a desirable solution to remedy this problem by leveraging previous experiences~\cite{braun2010structure}. A meta-learning model should be able to generalize a learning strategy to different tasks derived from a common distribution~\cite{thrun1998learning}. In contrast, traditional machine learning models are limited in adapting to a single task by finding patterns that generalize across data points.

Recently, several meta-learning methods have been proposed and demonstrated to be effective. The main objective is to provide a model for efficient learning of new tasks to help us get closer to Artificial General Intelligence (AGI). The primary idea behind meta-learning is to employ the acquired knowledge in solving previous tasks to generalize the learning to new tasks. In general, meta-learning algorithms can be classified into three types: metric-based, optimization-based, and model-based.

We can point to Matching Networks~\cite{vinyals2016matching}, Prototypical Networks~\cite{snell2017prototypical}, and Relation Networks~\cite{sung2018learning} as notable methods belonging to metric-based algorithms. They can allow unseen tasks to be efficiently learned while not experiencing catastrophic forgetting. However, they cannot be directly applied to other methods associated with reinforcement learning. Optimization-based algorithms are prevalent since they are model agnostic besides being task agnostic~\cite{finn2017model}. However, they are computationally inefficient since they are required to compute second-order gradients. MAML~\cite{finn2017model}, Meta-SGD~\cite{li2017meta}, and AVID~\cite{javaheri2022avid} are practical methods in this category. We can refer to MANN~\cite{santoro2016meta} and SNAIL~\cite{mishra2017simple} as noteworthy examples of model-based techniques. Most of these models include memory modules such as Recurrent Neural Networks (RNNs) or Long Short-Term Memory (LSTM) units, allowing them to store their experiences. Nevertheless, using these modules would increase the number of model parameters.

Despite being influenced by human brain behavior, most successful meta-learning approaches lack bio-plausibility. The human brain can quickly learn new skills by utilizing prior knowledge and experiences, which are mostly encoded in the temporal lobe of the human cortex~\cite{davachi2006item}. The temporal lobe is the second largest lobe, responsible for processing auditory information, language comprehension, and the formation of long-term memories~\cite{squire2004medial}. The hippocampus is one of the main regions in the medial temporal lobe, which plays a crucial role in episodic memory~\cite{eichenbaum2012towards}. As a type of declarative memory, episodic memory refers to the human ability to recall specific events and experiences from the past.

Wang et al.~\cite{wang2018prefrontal} have simulated the Prefrontal Cortex (PFC) behavior for meta-reinforcement learning using the assumption that the PFC also encodes the recent history of actions and rewards besides representing the expected values of actions, objects, and states. It has been conceived that PFC, along with the thalamic nuclei and basal ganglia, form a recurrent neural network~\cite{matsumoto2007medial,redish2007reconciling}. Through its inputs, this network is provided with information about the performed actions and received rewards. On the output side, actions and estimation of the state value are generated by the network~\cite{kennerley2009evaluating}. By incorporating PFC with dopamine (DA), Wang et al. have created a model that includes two full-fledged reinforcement learning systems, one based on activity representations and the other based on synaptic learning to develop a unique meta-learning algorithm~\cite{wang2018prefrontal}. The model was further enhanced by Ritter et al.~\cite{ritter2018been} with embedding the episodic recall mechanism to overcome catastrophic forgetting using differentiable-neural-dictionary (DND)~\cite{pritzel2017neural, kaiser2017learning} as long-term memory.

Although Wang et al.~\cite{wang2018prefrontal} and Ritter et al.~\cite{ritter2018been} have proposed models inspired by biological procedures, they use LSTM, making them to be less biologically plausible. However, using Spiking Neural Networks (SNNs) can provide a better solution to design biologically plausible models, as SNNs are more similar to the behavior of the human brain neural network~\cite{tavanaei2019deep}. In a sense, SNNs can be viewed as a particular case of RNNs with internal states similar to the LSTM~\cite{he2020comparing}. Also, SNNs are more energy-efficient than conventional artificial neural networks and can be easily incorporated into neuromorphic electronic systems~\cite{neftci2019surrogate}. Neuromorphic devices aim to capture the brain’s fundamental properties to enable low-power, versatile, and fast processing of information~\cite{mead1990neuromorphic,indiveri2011neuromorphic,davies2019benchmarks}. Stewart and Neftci~\cite{stewart2022meta} have utilized SNNs for meta-learning using surrogate gradient descent to take advantage of the MAML paradigm, which has achieved remarkable results. Subramoney et al.~\cite{subramoney2021revisiting} have investigated how synaptic plasticity and network dynamics contribute to fast learning in SNNs by allowing synaptic weights to store prior information via backpropagation through time (BPTT) to mimic the brain’s ability to learn from very few samples. Scherr et al.~\cite{scherr2020one} have proposed a model for one-shot learning using SNNs by simulating the ventral tegmental area (VTA) using an additional model to produce some learning signals as dopamine that enables fast learning via local synaptic plasticity in their network. However, all these models rely on gradient computations and backpropagation algorithms, which undermines the bio-plausibility of the model.

In this paper, we have utilized SNNs to simulate memory and decision-making systems using Spike-Timing-Dependent Plasticity (STDP)~\cite{gerstner1996neuronal,bi1998synaptic} and reward-modulated STDP (R-STDP)~\cite{fremaux2016neuromodulated,brzosko2017sequential} for synaptic plasticity. Using these methods, we propose a novel bio-plausible meta-learner based on the role of the hippocampus and the PFC to effectively solve new unseen tasks and prevent catastrophic forgetting by encoding past experiences into a memory layer. The required information for solving new tasks that share a similar structure with previously learned tasks can then be retrieved from this memory layer.

\vspace{7pt}
\section{Materials and Methods}
\label{sec:method}
We consider a few-shot learning task $\mathcal{T}$ to be sampled from an unknown task distribution $p(\mathcal{T})$. Each task is characterized by a distinct dataset $\mathcal{D} = \{(x_{i}, y_{i})  \, |   \, x_i \in X,\, y_i \in Y\}_{i=1}^{n}$, consisting of $n$ independent and identically distributed input-output pairs $(x_i, y_i)$.
The corresponding dataset of task $\mathcal{T}$ is divided into two sets: a support set $\mathcal{D}^\mathcal{S}$ and a query set $\mathcal{D}^\mathcal{Q}$, such that $\mathcal{D} = \mathcal{D}^\mathcal{S} \cup \mathcal{D}^\mathcal{Q}$.

\begin{figure}[ht]
\centering
\includegraphics[width=1\linewidth]{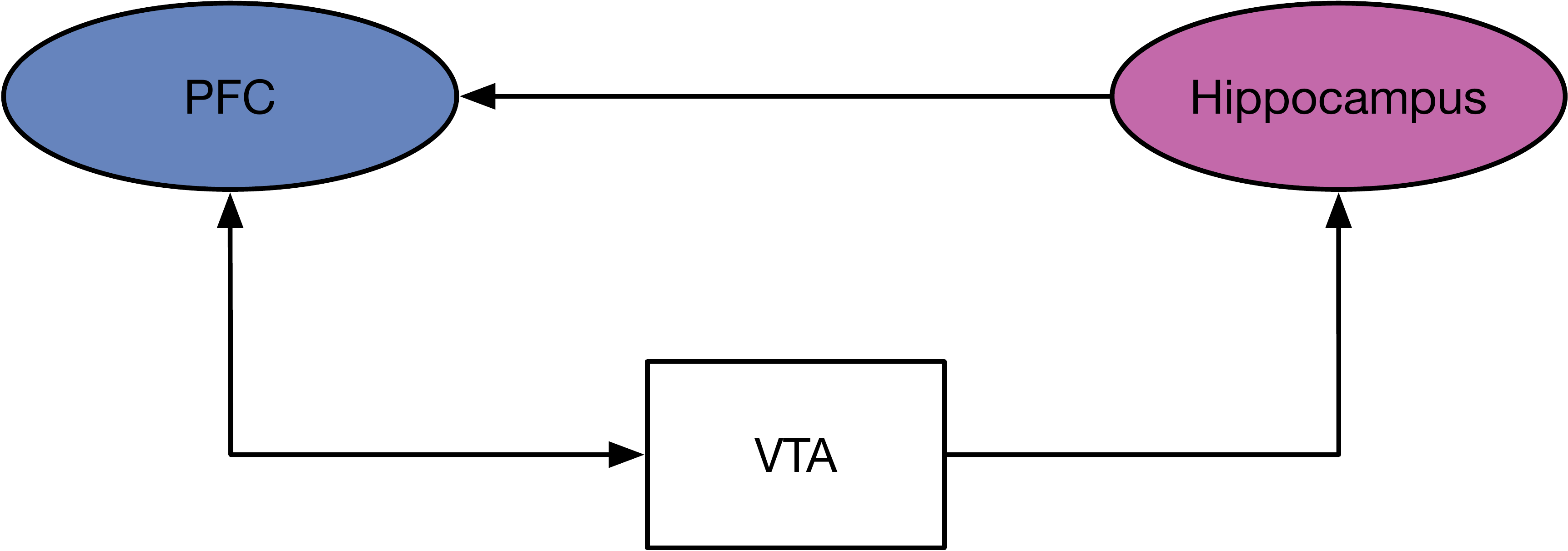}
\caption{The interactions between PFC, hippocampus, and VTA in the brain.}
\label{fig:vta}
\end{figure}

Here, we present a model to solve unseen tasks sampled from $p(\mathcal{T})$. The proposed model relies on the interactions between the hippocampus, PFC, and VTA, as shown in Figure~\ref{fig:vta}. Together, they create a memory system where the hippocampus memorizes long-term events for the future while the PFC receives signals from the hippocampus and other regions to make decisions. The PFC also includes short-term memory, or working memory, which can store recent events and accelerate decision-making~\cite{li2015hippocampal}.

Also, the ventral tegmental area (VTA) plays a crucial role in the brain's reward system. It contains dopaminergic neurons that release dopamine, which interacts with the prefrontal cortex (PFC) and hippocampus~\cite{li2015hippocampal}.

\vspace{7pt}

\begin{figure*}
\centering
\includegraphics[width=1\linewidth]{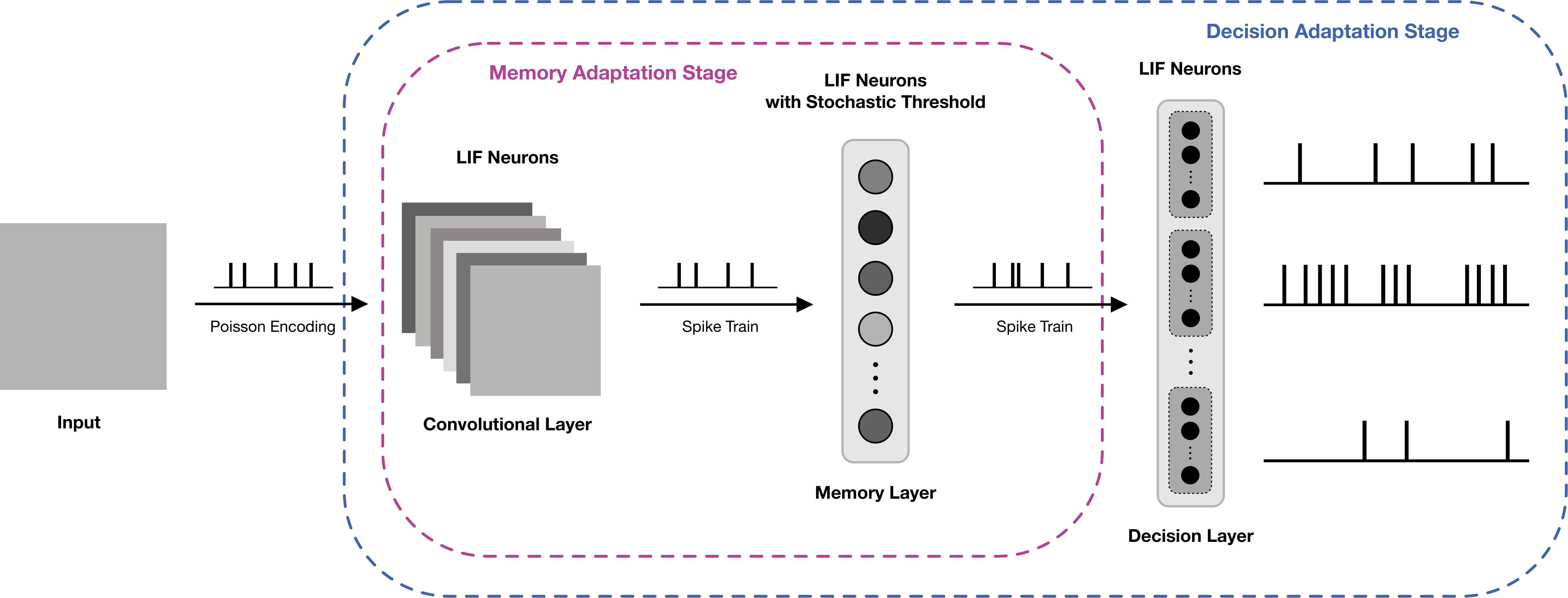}
\caption{The proposed model architecture overview.}
\label{fig:bio}
\end{figure*}

This work provides a new perspective on reward-based learning computations by considering bio-plausibility and simplicity as fundamental principles. With this in mind, we have utilized SNNs in order to enhance computational power in designing a highly dynamic meta-learner. Figure~\ref{fig:bio} depicts the overall architecture of the model, composed of three major components: the convolutional layer, the memory layer, and the decision layer. 
In this model, input data arrives at the convolutional layer, where it is processed and its features are extracted. The resulting features are then encoded in the memory layer, which serves as an episodic memory system for the model. Finally, the decision layer classifies the input data using the retrieved information from the memory layer, which provides the necessary context for accurate classification. 

Meta-learning includes two phases: the meta-training phase and the meta-testing phase. The meta-training phase involves performing both the memory and the decision adaptation stages. Accordingly, the memory layer is updated with the support set, whereas the decision layer is updated with the query set. During the meta-testing phase, the model is given a support set of few data points for each unseen task in order to update the memory layer through the memory adaptation stage. The updated memory layer is then utilized in conjunction with the decision layer to predict the labels of the corresponding query set.


\vspace{7pt}

\subsection{Convolutional Layer}
\label{subsec:conv}
Humans can recognize surrounding objects easily~\cite{thorpe1996speed} despite changes in object position, size, pose, illumination conditions, and background context~\cite{dicarlo2012does, biederman1987recognition}. Apparently, this invariant recognition is handled through a hierarchical process in the ventral pathway. It begins with the V1 area, which extracts simple features like bars and edges~\cite{lennie2005coding}. It continues through intermediate areas such as V2 and V4, which are responsive to more complex features~\cite{nandy2013fine}, and finally, in the inferior temporal cortex (IT), 
in which the neurons selectively respond to objects or parts of the objects~\cite{tanaka1991coding}.

In the same way as the ventral path, the convolutional layer extracts the features of the input image and feeds them to the next layer, i.e., the memory layer. The convolutional layer is composed of leaky integrate-and-fire (LIF) neurons, which receive Poisson-encoded images as inputs.

The LIF neuron model is a mathematical model used to describe the behavior of biological neurons, where the membrane potential depends on incoming excitatory and inhibitory inputs and passive ion leakage across the membrane. When the membrane potential reaches a certain threshold $u_{\theta}$, the neuron fires an action potential or spike, and the membrane potential is then reset to a lower value. The LIF model assumes that the membrane potential can be described by a single variable, $u(t)$, which represents the voltage across the neuron's membrane at time $t$. The equation governing the dynamics of $u(t)$ is as follows~\cite{gerstner2014neuronal}:
\begin{equation} 
    \label{eq:lif}
    \begin{split}
        \tau.\frac{du}{dt} = - (u - u_{rest}) + R.I(t),
    \end{split}
\end{equation}
where $u_{rest}$ denotes the voltage across the neuron's membrane in the rest mode, $R$ is the resistance of the neuron's membrane, and $I(t)$ represents its input current at time $t$. Also, $\tau$ is a time constant that corresponds to the leakage rate.

The training process of the convolutional layer is unsupervised and is performed through a particular form of synaptic plasticity known as spike-timing-dependent plasticity (STDP), which has also been observed to occur in the human visual cortex~\cite{mcmahon2012stimulus}. In general, STDP potentiates the afferent connections involved in making a neuron fire while depressing the others~\cite{feldman2012spike, kheradpisheh2018stdp}. 

We have utilized the standard STDP, where the weights are updated as follows:
\begin{equation}
\label{eq:stdp}
\Delta w_{ij} = 
\left\{ \begin{aligned} 
  & \,  A_{+} \exp{\frac{- \Delta t}{\tau_{+}}}, \;\; if \;\; \Delta t \geq 0,\\
  & \, -A_{-} \exp{\frac{\Delta t}{\tau_{-}}}, \;\; if \;\; \Delta t < 0,
\end{aligned} \right.
\end{equation}
where $i$ and $j$ respectively refer to the index of post- and pre-synaptic neurons, $\Delta t = t_{i} - t_{j}$, in which $t_{i}$ and $t_{j}$ are the corresponding spike times, and $\Delta w_{ij}$ is the synaptic weight change. Also, positive constants $A_{+}$ and $A_{-}$ scale the strength of potentiation and depression of weight changes, respectively, and $\tau_{+}$ and $\tau_{-}$ are positive time constants defining the width of the positive and negative learning window. 

In addition, we have utilized the soft bounding by multiplying the term $w_{ij}(1 - w_{ij})$ by $\Delta w_{ij}$ in order to keep the weights within the range $[0,1]$ and stabilize the weight changes as it converges~\cite{mozafari2019bio}.

\subsection{Memory Layer}
\label{subsec:memory}
Typically, human decisions are based on memories of previous events. It seems that we learn by storing summaries of individual episodes for long periods and then retrieving them when similar situations arise. Past experiences are recalled through episodic memory~\cite{eichenbaum2012towards}.

As part of meta-learning, the goal is to encode the information of past decisions into memories, which can be used to make future decisions more effectively. Specifically, the memory layer aims to mimic the hippocampus role and equip the model with an episodic memory system. 

Memories are made by changes in collections of neurons and the synaptic connections between them~\cite{jeong2021synaptic}. A memory may be encoded in one group of neural circuits and can be recalled in another. Every time a memory is recalled, it may change depending on the active neural circuits at that time~\cite{silva2009molecular}. When a memory is constantly recalled, its active connections become stronger.

Consequently, this layer encodes the information received from the convolutional layer within a portion of LIF neurons with stochastic thresholds, in which spikes of neuron $i$ are generated stochastically with stochastic intensity $\rho_i = g(u_i)$ in which $g(u_i)$ is an exponential function of the membrane potential \cite{jolivet2006predicting}:
\begin{equation} 
    \label{eq:threshold}
    \begin{split}
        \rho_{i} = g(u_i) = \rho_{\theta} exp\left(\frac{u_i - u_{\theta}}{\Delta u} \right),
    \end{split}
\end{equation}
where $\rho_{\theta}$ indicates the stochastic intensity at threshold, $u_{\theta}$ is the default firing threshold and $\Delta u$ defines the width of threshold region.

In the memory layer, LIF neurons with stochastic thresholds form an episodic memory system that can recall prior information to solve unseen tasks more effectively. The LIF neuron with stochastic threshold has been utilized to make the model robust for intrinsic or synaptic noises generated by pre-synaptic neurons to improve generalization.

The weights of the memory layer are updated during the memory adaptation stage, as indicated in Figure~\ref{fig:bio}. During this stage, the reward-modulated STDP (R-STDP) is used to ensure that large amounts of information can be encoded efficiently. Using R-STDP, each sample is forced to be encoded with a small subset of neurons. As a result, when previous experiences are recalled frequently, these neurons become more selective towards similar features extracted from the convolutional layer. This process enables the memory layer to adaptively learn and store relevant information.

R-STDP is a type of synaptic plasticity that is thought to underlie learning and memory in the brain~\cite{fremaux2016neuromodulated}. More precisely, it is a process by which the strength of synapses between neurons can be modified based on the relative timing of their activities, as well as reward signals provided by neuromodulators such as dopamine (DA)~\cite{fremaux2013reinforcement}. The reward signal typically comes from dopaminergic neurons in the midbrain, which release dopamine in response to rewarding or aversive stimuli~\cite{brzosko2017sequential}. R-STDP is a form of reinforcement learning that allows for a more nuanced and context-specific form of synaptic plasticity that takes into account the rewards and punishments associated with different patterns of neural activity~\cite{mozafari2019bio}. This process can refine and optimize neural circuits over time in response to environmental feedback, leading to enhance learning and adaptation. The R-STDP learning rule can be formulated by the following equations:
\begin{equation}
    \begin{split}
    \label{eq:rstdp}
        \frac{dc}{dt} &= -\frac{c}{\tau_{c}} + STDP(\Delta t) \delta (t - t_{pre/post}),\\ 
        \frac{dw}{dt} &= cd,\\
        \frac{dd}{dt} &= - \frac{d}{\tau_{d}} + DA(t),
    \end{split}
\end{equation} 
where $c$ represents the eligibility traces that act as synaptic tags, while $\Delta t = t_{post} - t_{pre}$ measures the time difference between post- and pre-synaptic activities. The variable $d$ describes the concentration of extracellular dopamine, and $\delta(t)$ is the Dirac delta function. Also, $\tau_c$ and $\tau_d$ are time constants of eligibility traces and DA uptake, respectively. Lastly, the function $DA(t)$ models the source of dopamine resulting from the activity of dopaminergic neurons in the midbrain.

In the memory adaptation stage, we aim to encode the extracted information of each sample in a portion of neurons of this layer using the R-STDP learning rule to maintain memory over time. It allows the network to adapt itself to new situations while retaining existing knowledge which improves its generalization capabilities and prevents catastrophic forgetting. Adjusting reward/punishment intervals allows only a certain percentage of neurons to fire; therefore, learning rarely occurs in synaptic weights. The network becomes more flexible and efficient when appropriately considering the reward/punishment intervals. The intervals along with the corresponding reward/punishment values are determined using the following piecewise function:
\begin{equation}
\label{eq:policy}
r(n_s) = 
\left\{ \begin{aligned} 
  & \, -2, \;\; if \;\; n_s < c-4s,\\
  & \, -1, \;\; if \;\; c-4s \leq n_s < c-2s,\\
  & \, +1, \;\; if \;\; c-2s \leq n_s < c-s,\\
  & \, +2, \;\; if \;\; c-s \leq n_s \leq c+s,\\
  & \, +1, \;\; if \;\; c+s < n_s \leq c+2s,\\
  & \, -1, \;\; if \;\; c+2s < n_s \leq c+4s,\\
  & \, -2, \;\; if \;\; c+4s < n_s,
\end{aligned} \right.
\end{equation}
where, $n_s$ is the percentage of activated neurons in this layer. Also, the sparsity level is specified by $c \pm s$, where $c$ is a constant that represents the average percentage of neurons that can fire, and $s$  is a spread percentage used to enhance the flexibility of the memory layer.


\subsection{Decision Layer}
\label{subsec:decision}
The prefrontal cortex (PFC) plays a crucial role in the brain, with decision-making being one of its most important functions~\cite{kim1999neural, shima1998role}. It receives input from multiple regions of the brain to appropriately process information as well as to make decisions, where the history of choices and rewards are dynamically encoded in PFC~\cite{seo2007dynamic, padoa2006neurons, tsutsui2016dynamic, seo2012action, barraclough2004prefrontal}. A reinforcement learning mechanism is evident in the neural activity of PFC~\cite{seo2008cortical}.

The principal objective of this layer in our model is to simulate the behavior of the PFC for decision-making using the reward-modulated STDP, introduced in Section~\ref{subsec:memory} as a reinforcement learning rule. This layer receives recalled information from its preceding layer, i.e., the memory layer, and processes it for decision-making. This layer consists of several groups of LIF neurons that are categorized based on the task. Each group of neurons includes $M$ neurons that can represent the input's class. 
Typically, for solving an $N$-way $K$-shot classification problem, this layer would have $N$ groups of neurons, each containing $M$ neurons, resulting in a total of $N \times M$ neurons. Each class could be represented by $M$ different neurons in a group, enhancing the generalization and versatility of the model. Finally, the input's class can be determined by the group of neurons that fire most frequently.

During the decision adaptation stage, this layer is trained by rewarding accurate predictions and punishing inaccurate ones. 
To update the weights more effectively, we have implemented adaptive R-STDP in this layer, in which the reward signal value is updated after solving each task to maintain a balance between rewards and punishments in the network. Initially, the reward and punishment values are $\nicefrac{1}{2}$ and $\nicefrac{-1}{2}$, respectively. These values are updated in each task containing $N$ samples by $\nicefrac{N_{incorrect}}{N}$ and $\nicefrac{N_{correct}}{N}$, respectively. Here, $N_{correct}$ and $N_{incorrect}$ represent the number of samples that are classified correctly and incorrectly over all samples of a task. Additionally, lateral inhibition mechanisms are utilized between different groups of neurons to make them compete. As a result, when the activity of one group of neurons is increased, the activities of other groups are suppressed, allowing the model to converge faster. Algorithm \ref{algo:meta} provides an outline of the complete algorithm in its general form.

\begin{algorithm}
\caption{Learning procedure for few-shot learning}\label{alg:cap}
\begin{algorithmic}[1]
\Require Task distribution $p(\mathcal{T})$ and corresponding dataset $\mathcal{D}$
\For{samples in $\mathcal{D}$}
    \State Compute $w_c$ for convolutional layer by Eq~\eqref{eq:stdp}
\EndFor
\State Randomly initialize $w_m$ and $w_d$ for memory and decision layers, respectively
\ForEach {epoch}
    \State Sample batch of tasks $\mathcal{T}_i = (\mathcal{D}^\mathcal{S}_i,\mathcal{D}^\mathcal{Q}_i)$ from $p(\mathcal{T})$
    \For{samples in $\mathcal{D}^\mathcal{S}_i$}
        \State Compute the percentage of activated neurons $(n_s)$
        \State Determine the corresponding reward/punishment to $n_s$, as described in Section~\ref{subsec:memory}
        \State Update $w_m$ by Eq~\eqref{eq:rstdp} 
    \EndFor
    
    \For{samples in $\mathcal{D}^\mathcal{Q}_i$}
        \State Determine the corresponding reward/punishment based on the decision layer's prediction, as described in Section~\ref{subsec:decision}
        \State Update $w_d$ by Eq~\eqref{eq:rstdp}
    \EndFor
\EndFor
\end{algorithmic}
\label{algo:meta}
\end{algorithm}

\vspace{7pt}
\section{Experiments}
\label{sec:experiments}
The performance of the proposed meta-learner is evaluated by testing its ability to solve various few-shot classification tasks. The primary purpose of this study was to determine whether this simple model could gain comparative results on few-shot learning benchmarks and whether it is capable of learning to learn.

\subsection{The Few-shot Learning Setting}


Meta-learning comprises two phases for training and evaluating a few-shot learning model: meta-training and meta-testing.

During the meta-training phase, we sample a batch of tasks $\mathcal{T}_{i}$ represented by $\mathcal{D}_{i}$, and the meta-learner is trained on the support set $\mathcal{D}_i^\mathcal{S}$ in the memory adaptation stage (task-level), followed by the decision adaptation stage (meta-level) on a query set $\mathcal{D}_i^\mathcal{Q}$ for each task.

In the meta-testing phase, the trained meta-learner is evaluated on a set of held-out unseen tasks $\mathcal{T}_u \sim p(\mathcal{T})$ that were not used for training. The model is given a support set $\mathcal{D}_u^\mathcal{S}$ corresponding to the new task $\mathcal{T}_u$ for the memory adaptation stage and then used to predict labels of the corresponding query set $\mathcal{D}_u^\mathcal{Q}$.

\subsection{Model Configuration}
\label{subsec:setup}
To set up the model for experiments, we first pre-trained the convolutional layer discussed in Section~\ref{subsec:conv} using random samples in the given dataset, where each sample is exposed to this network for $50ms$ through Poisson encoding. This layer contains $30$ convolutional filters with a kernel size of $8\times8$ and a stride size of $2$ for downsampling. To enhance the edge perception of the model, we also employed the lateral inhibition mechanism in this layer, where each neuron inhibits the activity of neighboring neurons of an individual filter using lateral connections.

In the memory layer, we used $N=100$ neurons with $u_{rest} = -70mV$, $\rho_{\theta} = 1/ms$, $u_{\theta} = -50mV$, and $\Delta u = 5mV$ in Equation~\eqref{eq:threshold}. Furthermore, we have included reward/punishment intervals from Section~\ref{subsec:memory} to regulate the sparsity of neuronal activity in the memory layer to around $15\%$ by using $c = 15\%$ and $s = 3\%$ in Equation~\eqref{eq:policy} to boost the model's generalization.


To make the decision layer scalable, we consider $M = 10$ neurons for each group representing a class in the decision layer, as discussed in Section~\ref{subsec:decision}.

\subsection{Few-Shot Classification Tasks}
The $N$-way $K$-shot classification problem involves $N$ different classes, each containing $K$ samples. A powerful meta-learner should be capable of recognizing inputs by comparing them rather than memorizing a definite mapping between those inputs and the desired classes. We scale up our approach to few-shot classification tasks using the Omniglot and Double MNIST datasets.

\begin{figure}[ht]
\centering
\includegraphics[width=1\linewidth]{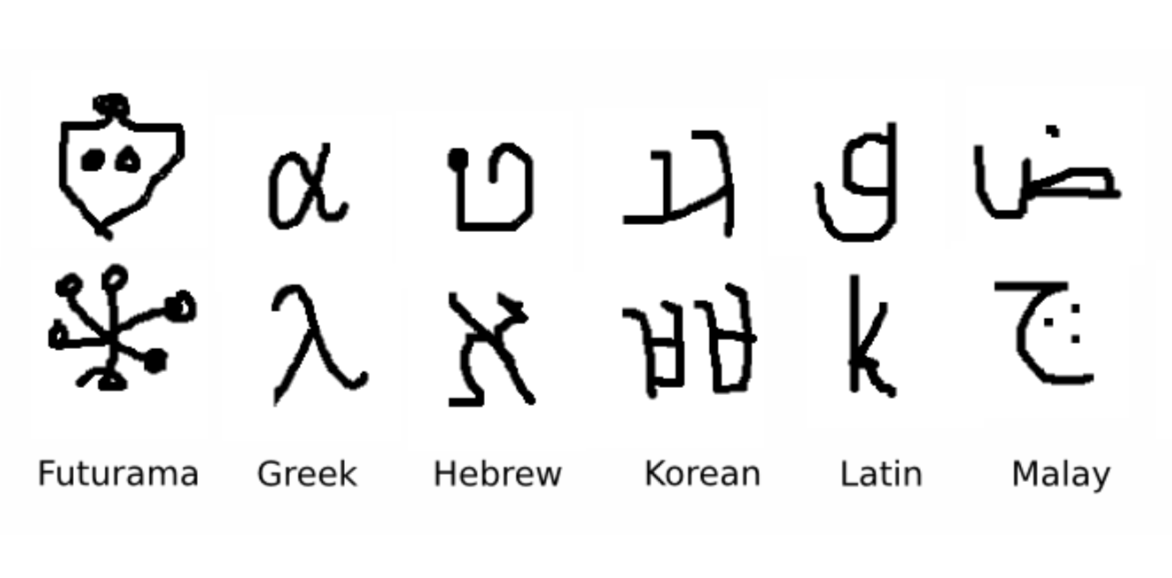}
\caption{Illustration of a few of the alphabets provided in Omniglot.}
\label{fig:omni}
\end{figure}

The Omniglot dataset is a well-known benchmark for few-shot learning introduced by Lake et al.~\cite{lake2019omniglot}.
This dataset includes handwritten characters from $50$ different languages, representing 1632 different classes, with $20$ samples in each class as shown in Figure~\ref{fig:omni}. According to the baseline models, 1200 characters are randomly selected for training, and the remaining ones are used for testing. We also applied data augmentation by rotating each instance of a class by a multiple of 90 degrees, following the approach suggested by Santoro et al.~\cite{santoro2016meta}.

\begin{figure}[ht]
\centering
\includegraphics[width=1\linewidth]{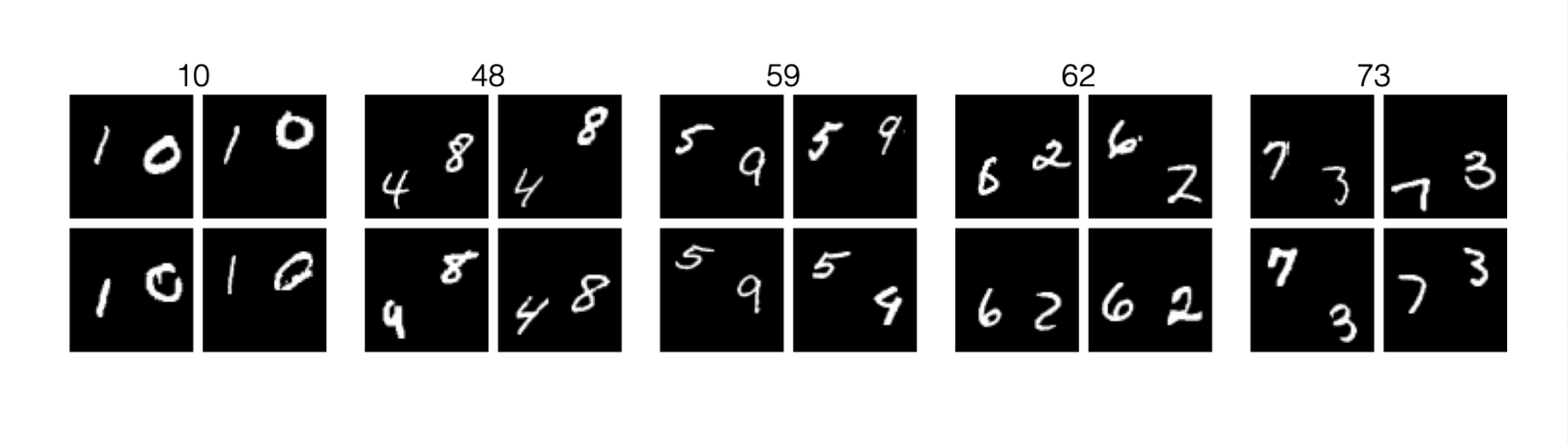}
\caption{Illustration of a few of the two-digit numbers provided in Double MNIST.}
\label{fig:dmnist}
\end{figure}

The Double MNIST dataset consists of 100 distinct classes of two-digit numbers, each containing 1000 unique handwritten samples as depicted in Figure~\ref{fig:dmnist}. For training purposes, 80 classes are randomly chosen, while the remaining 20 classes are preserved for testing.

In order to evaluate the model on the $N$-way $K$-shot classification problem, we feed $N \times K$ samples to the meta-learner in a random order to perform the memory adaptation stage explained in Section~\ref{subsec:memory}. We then present a new, unlabeled sample from one of the $N$ classes and report the average accuracy on this last $(N \times K+1)$-th pass based on the decision layer discussed in Section~\ref{subsec:decision} in a similar fashion to SNAIL~\cite{mishra2017simple}.

Using these standard datasets, we were able to demonstrate the efficiency and generalization capabilities of our proposed model.

\subsection{Results}
We present an analysis of the results obtained from solving few-shot classification tasks. Firstly, we examine the performance of our proposed model in solving 5-way 5-shot and 5-way 1-shot classification tasks on the Omniglot dataset, as shown in Table~\ref{table:omni}.

Our proposed model outperformed other models such as Siamese Networks~\cite{koch2015siamese}, Matching Networks~\cite{vinyals2016matching}, Prototypical Networks~\cite{snell2017prototypical}, and Meta Networks~\cite{munkhdalai2017meta} in solving 5-way 5-shot tasks on Omniglot. It also performed similarly to the state-of-the-art meta-learning models, namely MAML~\cite{finn2017meta} and SNAIL~\cite{mishra2017simple}. Moreover, our proposed model achieved superior results in solving 5-way 1-shot tasks on the same dataset compared to the other competitors. 

By comparing the results of our proposed model with those of comparable simple and generic methods such as MANN~\cite{santoro2016meta}, it can be concluded that our proposed model is highly effective in solving problems with a small number of samples.

MANN~\cite{santoro2016meta} is a regular method for meta-learning that utilizes Neural Turing Machine (NTM)~\cite{graves2014neural} as the embedded external memory. This model stores the information of each task in its memory so that when it encounters a new one, it recovers the information needed to solve that task by measuring its cosine similarity with the information stored in the memory. The disadvantage of this method is that it requires more memory than our model since our proposed model uses different combinations of neurons of the episodic memory layer to simulate memory. Moreover, the episodic memory layer in our model is more flexible than the embedded memory in MANN~\cite{santoro2016meta}. While MANN retrieves information using cosine similarity, which returns a row of information related to a task that has the highest degree of similarity, some tasks in practice are composed of a combination of several tasks and cannot be solved by retrieving information from one previously solved task. Our proposed model overcomes this limitation by providing an episodic memory system that resembles the human brain.

\begin{table}[ht]
\renewcommand{\arraystretch}{1.3}
\caption{Comparison of test accuracies by our proposed model and other methods on the Omniglot dataset.}
\label{table:omni}
\centering
\begin{tabular}{c||c|c}
\hline
Method & \multicolumn{2}{c}{5-way Omniglot}  \\ 
\hline
 & 1-shot & 5-shot\\
\hline
MANN~\cite{santoro2016meta} & $82.8\%$  & $94.9\%$\\
Siamese Networks~\cite{koch2015siamese} & $97.3\%$ & $98.4\%$\\
Matching Networks~\cite{vinyals2016matching} & $98.1\%$ & $98.9\%$\\
MAML~\cite{finn2017meta} & $98.7 \pm 0.4\%$ & \bm{$99.9 \pm 0.3\%$}\\
Prototypical Networks~\cite{snell2017prototypical} & $97.4\%$ & $99.3\%$\\
Meta Networks~\cite{munkhdalai2017meta} & $98.9\%$ & -\\
SNAIL~\cite{mishra2017simple} & $99.07 \pm 0.16\%$ & $99.78 \pm 0.09\%$\\
\hline
Ours & \bm{$99.06 \pm 0.24\%$} & $99.53 \pm 0.15\%$\\
\hline
\end{tabular}
\end{table}

In another experiment, we evaluated the performance of our model in solving $5$-way $1$-shot tasks on the Double MNIST dataset, with results presented in Table~\ref{table:dmnist}. Our primary objective was to compare our model's performance with a previously introduced meta-learning model, which is essentially the same as MAML and its first-order variation (FOMAML) but uses spiking neurons as its building blocks with surrogate gradient descent as the learning rule~\cite{stewart2022meta}.
In spite of their important work embedding SNNs into MAML, it was expected to have the same or weaker performance than the original MAML model, as confirmed by the results.
However, our proposed model is entirely different in terms of network design and learning rules, and we attempted to provide a new meta-learning approach that is consistent with biological observations.

The results show that our model can compete with other meta-learning models, despite being based on biologically plausible learning rules that are less computationally complex than the surrogate gradient descent~\cite{neftci2019surrogate}. It should be noted that previous models were all based on gradient computations, which are computationally expensive and not biologically plausible. Our model takes advantage of the potential of SNNs in meta-learning, which is difficult to implement using conventional artificial neural networks.

\begin{table}[ht]
\renewcommand{\arraystretch}{1.3}
\caption{Comparison of test accuracies by our proposed model and other methods on the Double MNIST dataset.}
\label{table:dmnist}
\centering
\begin{tabular}{c||c}
\hline
Method & 5-way 1-shot Double MNIST  \\ 
\hline
MAML (CNN)~\cite{finn2017meta} & \bm{$98.35 \pm 1.26\%$} \\
MAML (SNN)~\cite{neftci2019surrogate} & $98.23 \pm 1.12\%$  \\
FOMAML (SNN)~\cite{neftci2019surrogate} & $92.63 \pm 0.74\%$ \\
\hline
Ours & $98.52 \pm 0.23\%$ \\
\hline
\end{tabular}
\end{table}

We further analyze the effect of the sparsity level in the memory layer to understand its impact when we encode the information of each sample in either a smaller or larger portion of neurons by considering the appropriate reward/ punishment intervals. Table~\ref{table:sparsity} presents the results corresponding to different sparsity levels in solving 5-way 1-shot tasks using the Omniglot and Double MNIST datasets. It shows that encoding information in too few neurons may not yield the best performance as it lacks sufficient information for decision-making. Conversely, encoding information in a large portion of neurons can also decrease the model's performance as it cannot efficiently preserve previously learned information. 
By comparing the results of different sparsity levels for Omniglot (5-way 1-shot) and Double MNIST (5-way 1-shot), we can observe that decreasing the sparsity level of the memory layer reduces the performance of the model in both tasks. However, this effect is more pronounced for Omniglot, which contains more classes than Double MNIST. Due to this, it is more susceptible to catastrophic forgetting when compared to Double MNIST.
Therefore, selecting an appropriate sparsity level for the memory layer is crucial to achieve high performance and maintain the knowledge learned from previous tasks to prevent catastrophic forgetting.

\begin{table}[ht]
\renewcommand{\arraystretch}{1.3}
\caption{Comparison of different sparsity levels in the memory layer of our proposed model.}
\label{table:sparsity}
\centering
\begin{tabular}{c||c|c}
\hline
Sparsity Level & 5-way 1-shot Omniglot & 5-way 1-shot Double MNIST \\ 
\hline
$5 \pm 1\%$ & $96.53 \pm 0.34\%$ & $96.74 \pm 0.39\%$ \\
$10 \pm 2\%$ & $98.46 \pm 0.37\%$ & $98.09 \pm 0.29\%$ \\
$15 \pm 3\%$ & \bm{$99.06 \pm 0.24\%$} & \bm{$98.52 \pm 0.23\%$} \\
$20 \pm 4\%$ & $97.23 \pm 0.35\%$ & $97.17 \pm 0.20\%$ \\ 
$40 \pm 8\%$ & $89.04 \pm 0.80\%$ & $96.28 \pm 0.37\%$ \\ 
\hline
\end{tabular}
\end{table}

\vspace{7pt}
\section{Discussion}
\label{sec:discussion}
Meta-learning algorithms have recently attracted the attention of many researchers due to their ability to learn new tasks with just a few samples. Despite the remarkable performance of existing meta-learning models, designing a lightweight and efficient bio-plausible meta-learning model is a critical challenge. Bio-plausible neural networks must have three essential characteristics~\cite{hao2020biologically}: (i) the ability to integrate temporal input and generate spikes, (ii) the use of spike-based computation for both training and inference, (iii) and the ability to use learning rules based on findings from biological experiments. As mentioned in Section~\ref{sec:intro}, some biologically inspired models have been proposed. Wang et al.~\cite{wang2018prefrontal} have discovered the role of PFC as a meta-reinforcement learning system. Later on, Ritter et al.~\cite{ritter2018been} improved the prior work and utilized the differentiable neural dictionary (DND) to store some information to solve new tasks. Although both models are the basis for our work, they use LSTMs, DND, gradient descent, and backpropagation, hindering their bio-plausibility. Stewart and Neftci have embedded SNNs into MAML, which is highly advantageous. Still, they use surrogate gradient and backpropagation, restricting their algorithm's bio-plausibility.
Furthermore, Subramoney et al.~\cite{subramoney2021revisiting}, and Scherr et al.~\cite{scherr2020one} have proposed a novel approach to enable quick learning in SNNs; however, their models do not meet bio-plausibility requirements. Here, we used LIF neurons along with their stochastic threshold variation incorporating a combination of STDP and R-STDP learning rules. Accordingly, the proposed model is highly bio-plausible as it meets all the aforementioned requirements.

On the other hand, the efficiency and hardware friendliness of SNNs makes them an appropriate choice for neuromorphic hardware deployment. Neuromorphic hardware is particularly well suited for online learning at the edge. In spite of this, they face several challenges, such as learning from scratch on data-hungry models due to robustness and time-to-convergence issues. Here, we have presented a lightweight and simple SNN model with a pre-trained convolutional layer that can alleviate these issues. According to the results, our model can learn new tasks in a few-shot setting. As a result, this enables learning in low-data regimes since accessing sufficient labeled data is demanding. In addition, our proposed model features a highly efficient episodic memory that significantly mitigates the issue of catastrophic forgetting. This is achieved through the memory's ability to store a vast amount of information and link similar underlying patterns, received from the convolutional layer. Consequently, it can effectively and partially recall previously stored data when presented with new, similar data. As a result of this functionality, the proposed model exhibits a high degree of generalization, enabling it to adapt to novel tasks with few examples.

\begin{figure}[ht]
\centering
\includegraphics[width=0.95\linewidth]{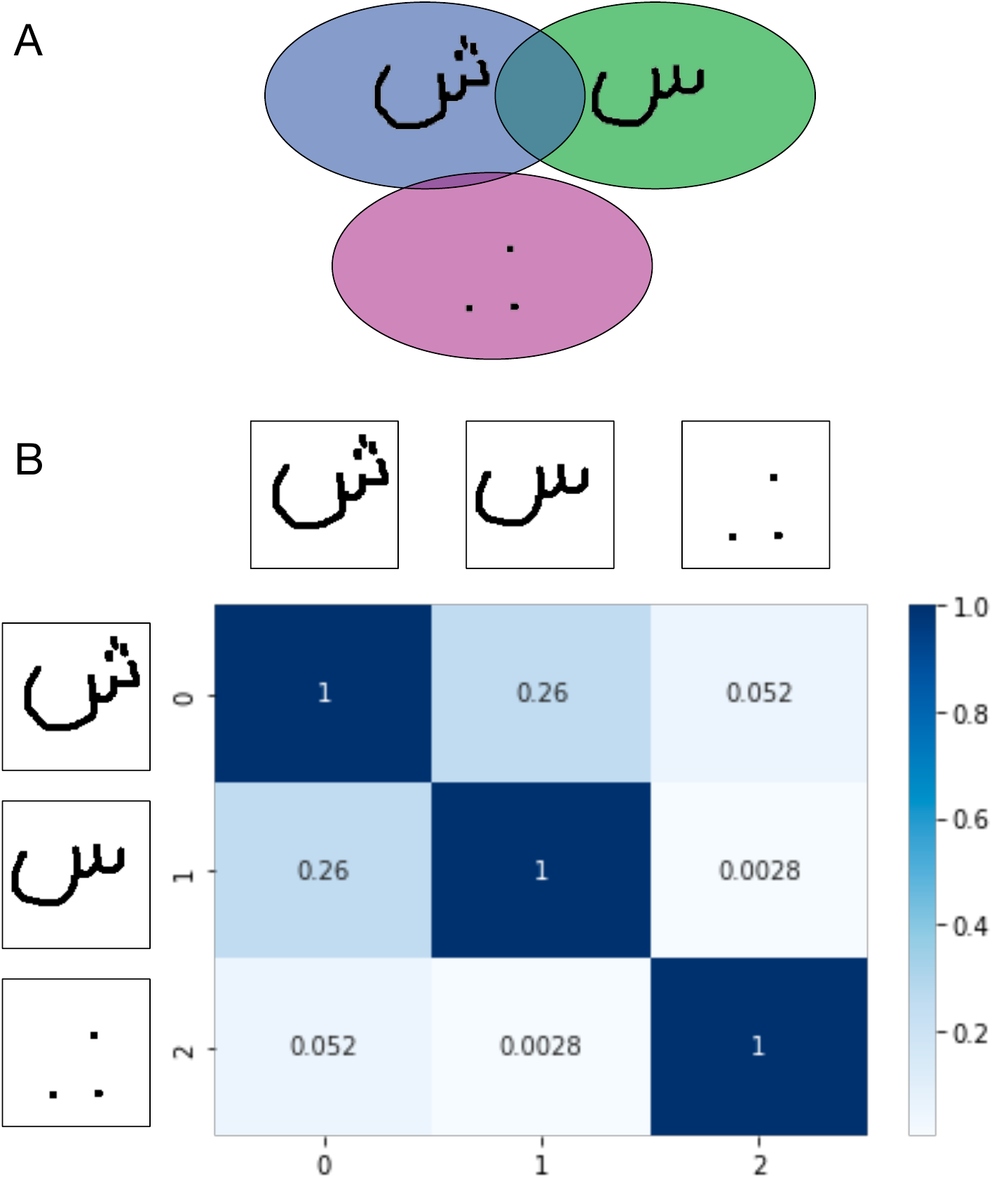}
\caption{A. Memory layer representation space. B. Comparison of memory representations of three different examples of the Omniglot dataset using the Pearson correlation coefficient.}
\label{fig:heatmap}
\end{figure}

We conducted an analysis to assess the efficacy of the memory layer, which involved the memory representation of a subset of samples from the Omniglot dataset, as depicted in Figure~\ref{fig:heatmap}. The memory representation of each sample is a binary vector that indicates which neurons were activated for that sample. To evaluate the correlation between pairs of samples, we calculated their Pearson correlation coefficient and generated a corresponding heatmap, which is also presented in Figure~\ref{fig:heatmap}. Higher values on the heatmap indicate a stronger correlation between the memory representations of the respective samples. Our results demonstrate that the sample with index 0 shares similar features with sample indices 1 and 2, resulting in a high correlation with sample index 1 due to their high degree of similarity, as well as a correlation with sample index 2 given the presence of some shared patterns. In contrast, sample index 1 has a low correlation with sample index 2 since they do not share any analogous patterns. These findings suggest that the memory layer functions appropriately by partially selecting neurons for underlying similarities.

\vspace{7pt}

\vspace{7pt}
\section{Conclusion and Future Work}
\label{sec:conclusion}
We presented a bio-plausible meta-learner with spiking neural networks (SNNs), motivated by the need for a meta-learner to be capable of mimicking the human brain's behavior when encountering a new unseen task. We have appropriately implemented the episodic memory and decision-making system by studying the role of the Hippocampus and the Prefrontal Cortex (PFC) to enable quick learning by incorporating past experiences. Our proposed model can be considered as an adaptive lifelong learning algorithm because it can attend to experiences over a lifetime and overcome catastrophic forgetting. As a result, our model can learn faster and generalize better due to its lifelong memory. Furthermore, it is computationally very efficient as it relies on spike-timing-dependent plasticity (STDP) that empowers the memory layer to decide what experiences are worth remembering. 

Our work represents an initial step toward developing a bio-plausible meta-learning model using SNNs, and future works could build on this foundation to address more complex tasks. Moreover, by scaling these ideas, we could facilitate the implementation of meta-learning models on spike-based neuromorphic devices, enabling fast learning on neuromorphic hardware.

\vspace{7pt}

\bibliographystyle{elsarticle-num}
\bibliography{biblio}

\begin{thebibliography}{10}
\expandafter\ifx\csname url\endcsname\relax
  \def\url#1{\texttt{#1}}\fi
\expandafter\ifx\csname urlprefix\endcsname\relax\def\urlprefix{URL }\fi
\expandafter\ifx\csname href\endcsname\relax
  \def\href#1#2{#2} \def\path#1{#1}\fi

\bibitem{schmidhuber2015deep}
J.~Schmidhuber, Deep learning in neural networks: An overview, Neural networks
  61 (2015) 85--117.

\bibitem{braun2010structure}
D.~A. Braun, C.~Mehring, D.~M. Wolpert, Structure learning in action,
  Behavioural brain research 206~(2) (2010) 157--165.

\bibitem{thrun1998learning}
S.~Thrun, L.~Pratt, Learning to learn: Introduction and overview, in: Learning
  to learn, Springer, 1998, pp. 3--17.

\bibitem{vinyals2016matching}
O.~Vinyals, C.~Blundell, T.~Lillicrap, D.~Wierstra, et~al., Matching networks
  for one shot learning, Advances in neural information processing systems 29
  (2016).

\bibitem{snell2017prototypical}
J.~Snell, K.~Swersky, R.~Zemel, Prototypical networks for few-shot learning,
  Advances in neural information processing systems 30 (2017).

\bibitem{sung2018learning}
F.~Sung, Y.~Yang, L.~Zhang, T.~Xiang, P.~H. Torr, T.~M. Hospedales, Learning to
  compare: Relation network for few-shot learning, in: Proceedings of the IEEE
  conference on computer vision and pattern recognition, 2018, pp. 1199--1208.

\bibitem{finn2017model}
C.~Finn, P.~Abbeel, S.~Levine, Model-agnostic meta-learning for fast adaptation
  of deep networks, in: International conference on machine learning, PMLR,
  2017, pp. 1126--1135.

\bibitem{li2017meta}
Z.~Li, F.~Zhou, F.~Chen, H.~Li, Meta-sgd: Learning to learn quickly for
  few-shot learning, arXiv preprint arXiv:1707.09835 (2017).

\bibitem{javaheri2022avid}
A.~Javaheri, A.~Gholamzadeh~Khoee, S.~R. Kheradpisheh, H.~Farahani,
  M.~Ganjtabesh, Avid: A variational inference deliberation for meta-learning,
  in: 2022 12th International Conference on Computer and Knowledge Engineering
  (ICCKE), IEEE, 2022, pp. 268--273.

\bibitem{santoro2016meta}
A.~Santoro, S.~Bartunov, M.~Botvinick, D.~Wierstra, T.~Lillicrap, Meta-learning
  with memory-augmented neural networks, in: International conference on
  machine learning, PMLR, 2016, pp. 1842--1850.

\bibitem{mishra2017simple}
N.~Mishra, M.~Rohaninejad, X.~Chen, P.~Abbeel, A simple neural attentive
  meta-learner, arXiv preprint arXiv:1707.03141 (2017).

\bibitem{davachi2006item}
L.~Davachi, Item, context and relational episodic encoding in humans, Current
  opinion in neurobiology 16~(6) (2006) 693--700.

\bibitem{squire2004medial}
L.~R. Squire, C.~E. Stark, R.~E. Clark, The medial temporal lobe, Annu. Rev.
  Neurosci. 27 (2004) 279--306.

\bibitem{eichenbaum2012towards}
H.~Eichenbaum, M.~Sauvage, N.~Fortin, R.~Komorowski, P.~Lipton, Towards a
  functional organization of episodic memory in the medial temporal lobe,
  Neuroscience \& Biobehavioral Reviews 36~(7) (2012) 1597--1608.

\bibitem{wang2018prefrontal}
J.~X. Wang, Z.~Kurth-Nelson, D.~Kumaran, D.~Tirumala, H.~Soyer, J.~Z. Leibo,
  D.~Hassabis, M.~Botvinick, Prefrontal cortex as a meta-reinforcement learning
  system, Nature neuroscience 21~(6) (2018) 860--868.

\bibitem{matsumoto2007medial}
M.~Matsumoto, K.~Matsumoto, H.~Abe, K.~Tanaka, Medial prefrontal cell activity
  signaling prediction errors of action values, Nature neuroscience 10~(5)
  (2007) 647--656.

\bibitem{redish2007reconciling}
A.~D. Redish, S.~Jensen, A.~Johnson, Z.~Kurth-Nelson, Reconciling reinforcement
  learning models with behavioral extinction and renewal: implications for
  addiction, relapse, and problem gambling., Psychological review 114~(3)
  (2007) 784.

\bibitem{kennerley2009evaluating}
S.~W. Kennerley, J.~D. Wallis, Evaluating choices by single neurons in the
  frontal lobe: outcome value encoded across multiple decision variables,
  European Journal of Neuroscience 29~(10) (2009) 2061--2073.

\bibitem{ritter2018been}
S.~Ritter, J.~Wang, Z.~Kurth-Nelson, S.~Jayakumar, C.~Blundell, R.~Pascanu,
  M.~Botvinick, Been there, done that: Meta-learning with episodic recall, in:
  International conference on machine learning, PMLR, 2018, pp. 4354--4363.

\bibitem{pritzel2017neural}
A.~Pritzel, B.~Uria, S.~Srinivasan, A.~P. Badia, O.~Vinyals, D.~Hassabis,
  D.~Wierstra, C.~Blundell, Neural episodic control, in: International
  Conference on Machine Learning, PMLR, 2017, pp. 2827--2836.

\bibitem{kaiser2017learning}
{\L}.~Kaiser, O.~Nachum, A.~Roy, S.~Bengio, Learning to remember rare events,
  arXiv preprint arXiv:1703.03129 (2017).

\bibitem{tavanaei2019deep}
A.~Tavanaei, M.~Ghodrati, S.~R. Kheradpisheh, T.~Masquelier, A.~Maida, Deep
  learning in spiking neural networks, Neural networks 111 (2019) 47--63.

\bibitem{he2020comparing}
W.~He, Y.~Wu, L.~Deng, G.~Li, H.~Wang, Y.~Tian, W.~Ding, W.~Wang, Y.~Xie,
  Comparing snns and rnns on neuromorphic vision datasets: Similarities and
  differences, Neural Networks 132 (2020) 108--120.

\bibitem{neftci2019surrogate}
E.~O. Neftci, H.~Mostafa, F.~Zenke, Surrogate gradient learning in spiking
  neural networks: Bringing the power of gradient-based optimization to spiking
  neural networks, IEEE Signal Processing Magazine 36~(6) (2019) 51--63.

\bibitem{mead1990neuromorphic}
C.~Mead, Neuromorphic electronic systems, Proceedings of the IEEE 78~(10)
  (1990) 1629--1636.

\bibitem{indiveri2011neuromorphic}
G.~Indiveri, B.~Linares-Barranco, T.~J. Hamilton, A.~v. Schaik,
  R.~Etienne-Cummings, T.~Delbruck, S.-C. Liu, P.~Dudek, P.~H{\"a}fliger,
  S.~Renaud, et~al., Neuromorphic silicon neuron circuits, Frontiers in
  neuroscience 5 (2011) 73.

\bibitem{davies2019benchmarks}
M.~Davies, Benchmarks for progress in neuromorphic computing, Nature Machine
  Intelligence 1~(9) (2019) 386--388.

\bibitem{stewart2022meta}
K.~M. Stewart, E.~O. Neftci, Meta-learning spiking neural networks with
  surrogate gradient descent, Neuromorphic Computing and Engineering 2~(4)
  (2022) 044002.

\bibitem{subramoney2021revisiting}
A.~Subramoney, G.~Bellec, F.~Scherr, R.~Legenstein, W.~Maass, Revisiting the
  role of synaptic plasticity and network dynamics for fast learning in spiking
  neural networks, bioRxiv (2021).

\bibitem{scherr2020one}
F.~Scherr, C.~St{\"o}ckl, W.~Maass, One-shot learning with spiking neural
  networks, BioRxiv (2020).

\bibitem{gerstner1996neuronal}
W.~Gerstner, R.~Kempter, J.~L. Van~Hemmen, H.~Wagner, A neuronal learning rule
  for sub-millisecond temporal coding, Nature 383~(6595) (1996) 76--78.

\bibitem{bi1998synaptic}
G.-q. Bi, M.-m. Poo, Synaptic modifications in cultured hippocampal neurons:
  dependence on spike timing, synaptic strength, and postsynaptic cell type,
  Journal of neuroscience 18~(24) (1998) 10464--10472.

\bibitem{fremaux2016neuromodulated}
N.~Fr{\'e}maux, W.~Gerstner, Neuromodulated spike-timing-dependent plasticity,
  and theory of three-factor learning rules, Frontiers in neural circuits 9
  (2016) 85.

\bibitem{brzosko2017sequential}
Z.~Brzosko, S.~Zannone, W.~Schultz, C.~Clopath, O.~Paulsen, Sequential
  neuromodulation of hebbian plasticity offers mechanism for effective
  reward-based navigation, Elife 6 (2017) e27756.

\bibitem{li2015hippocampal}
M.~Li, C.~Long, L.~Yang, Hippocampal-prefrontal circuit and disrupted
  functional connectivity in psychiatric and neurodegenerative disorders,
  BioMed research international 2015 (2015).

\bibitem{thorpe1996speed}
S.~Thorpe, D.~Fize, C.~Marlot, Speed of processing in the human visual system,
  nature 381~(6582) (1996) 520--522.

\bibitem{dicarlo2012does}
J.~J. DiCarlo, D.~Zoccolan, N.~C. Rust, How does the brain solve visual object
  recognition?, Neuron 73~(3) (2012) 415--434.

\bibitem{biederman1987recognition}
I.~Biederman, Recognition-by-components: a theory of human image
  understanding., Psychological review 94~(2) (1987) 115.

\bibitem{lennie2005coding}
P.~Lennie, J.~A. Movshon, Coding of color and form in the geniculostriate
  visual pathway (invited review), JOSA A 22~(10) (2005) 2013--2033.

\bibitem{nandy2013fine}
A.~S. Nandy, T.~O. Sharpee, J.~H. Reynolds, J.~F. Mitchell, The fine structure
  of shape tuning in area v4, Neuron 78~(6) (2013) 1102--1115.

\bibitem{tanaka1991coding}
K.~Tanaka, H.-a. Saito, Y.~Fukada, M.~Moriya, Coding visual images of objects
  in the inferotemporal cortex of the macaque monkey, Journal of
  neurophysiology 66~(1) (1991) 170--189.

\bibitem{gerstner2014neuronal}
W.~Gerstner, W.~M. Kistler, R.~Naud, L.~Paninski, Neuronal dynamics: From
  single neurons to networks and models of cognition, Cambridge University
  Press, 2014.

\bibitem{mcmahon2012stimulus}
D.~B. McMahon, D.~A. Leopold, Stimulus timing-dependent plasticity in
  high-level vision, Current biology 22~(4) (2012) 332--337.

\bibitem{feldman2012spike}
D.~E. Feldman, The spike-timing dependence of plasticity, Neuron 75~(4) (2012)
  556--571.

\bibitem{kheradpisheh2018stdp}
S.~R. Kheradpisheh, M.~Ganjtabesh, S.~J. Thorpe, T.~Masquelier, Stdp-based
  spiking deep convolutional neural networks for object recognition, Neural
  Networks 99 (2018) 56--67.

\bibitem{mozafari2019bio}
M.~Mozafari, M.~Ganjtabesh, A.~Nowzari-Dalini, S.~J. Thorpe, T.~Masquelier,
  Bio-inspired digit recognition using reward-modulated spike-timing-dependent
  plasticity in deep convolutional networks, Pattern recognition 94 (2019)
  87--95.

\bibitem{jeong2021synaptic}
Y.~Jeong, H.-Y. Cho, M.~Kim, J.-P. Oh, M.~S. Kang, M.~Yoo, H.-S. Lee, J.-H.
  Han, Synaptic plasticity-dependent competition rule influences memory
  formation, Nature communications 12~(1) (2021) 3915.

\bibitem{silva2009molecular}
A.~J. Silva, Y.~Zhou, T.~Rogerson, J.~Shobe, J.~Balaji, Molecular and cellular
  approaches to memory allocation in neural circuits, Science 326~(5951) (2009)
  391--395.

\bibitem{jolivet2006predicting}
R.~Jolivet, A.~Rauch, H.-R. L{\"u}scher, W.~Gerstner, Predicting spike timing
  of neocortical pyramidal neurons by simple threshold models, Journal of
  computational neuroscience 21 (2006) 35--49.

\bibitem{fremaux2013reinforcement}
N.~Fr{\'e}maux, H.~Sprekeler, W.~Gerstner, Reinforcement learning using a
  continuous time actor-critic framework with spiking neurons, PLoS
  computational biology 9~(4) (2013) e1003024.

\bibitem{kim1999neural}
J.-N. Kim, M.~N. Shadlen, Neural correlates of a decision in the dorsolateral
  prefrontal cortex of the macaque, Nature neuroscience 2~(2) (1999) 176--185.

\bibitem{shima1998role}
K.~Shima, J.~Tanji, Role for cingulate motor area cells in voluntary movement
  selection based on reward, Science 282~(5392) (1998) 1335--1338.

\bibitem{seo2007dynamic}
H.~Seo, D.~J. Barraclough, D.~Lee, Dynamic signals related to choices and
  outcomes in the dorsolateral prefrontal cortex, Cerebral Cortex 17~(suppl\_1)
  (2007) i110--i117.

\bibitem{padoa2006neurons}
C.~Padoa-Schioppa, J.~A. Assad, Neurons in the orbitofrontal cortex encode
  economic value, Nature 441~(7090) (2006) 223--226.

\bibitem{tsutsui2016dynamic}
K.-I. Tsutsui, F.~Grabenhorst, S.~Kobayashi, W.~Schultz, A dynamic code for
  economic object valuation in prefrontal cortex neurons, Nature communications
  7~(1) (2016) 12554.

\bibitem{seo2012action}
M.~Seo, E.~Lee, B.~B. Averbeck, Action selection and action value in
  frontal-striatal circuits, Neuron 74~(5) (2012) 947--960.

\bibitem{barraclough2004prefrontal}
D.~J. Barraclough, M.~L. Conroy, D.~Lee, Prefrontal cortex and decision making
  in a mixed-strategy game, Nature neuroscience 7~(4) (2004) 404--410.

\bibitem{seo2008cortical}
H.~Seo, D.~Lee, Cortical mechanisms for reinforcement learning in competitive
  games, Philosophical Transactions of the Royal Society B: Biological Sciences
  363~(1511) (2008) 3845--3857.

\bibitem{lake2019omniglot}
B.~M. Lake, R.~Salakhutdinov, J.~B. Tenenbaum, The omniglot challenge: a 3-year
  progress report, Current Opinion in Behavioral Sciences 29 (2019) 97--104.

\bibitem{koch2015siamese}
G.~Koch, R.~Zemel, R.~Salakhutdinov, et~al., Siamese neural networks for
  one-shot image recognition, in: ICML deep learning workshop, Vol.~2, Lille,
  2015.

\bibitem{munkhdalai2017meta}
T.~Munkhdalai, H.~Yu, Meta networks, in: International conference on machine
  learning, PMLR, 2017, pp. 2554--2563.

\bibitem{finn2017meta}
C.~Finn, S.~Levine, Meta-learning and universality: Deep representations and
  gradient descent can approximate any learning algorithm, arXiv preprint
  arXiv:1710.11622 (2017).

\bibitem{graves2014neural}
A.~Graves, G.~Wayne, I.~Danihelka, Neural turing machines, arXiv preprint
  arXiv:1410.5401 (2014).

\bibitem{hao2020biologically}
Y.~Hao, X.~Huang, M.~Dong, B.~Xu, A biologically plausible supervised learning
  method for spiking neural networks using the symmetric stdp rule, Neural
  Networks 121 (2020) 387--395.

\end{thebibliography}

\end{document}